%% file: main.tex
\documentclass[10pt,twocolumn,letterpaper]{article}

\usepackage[pagenumbers]{cvpr} % To force page numbers, e.g. for an arXiv version

\input{preamble}

\usepackage{algorithm}
\usepackage{algpseudocode}
\usepackage{booktabs}
\usepackage{multirow}
\usepackage{amssymb}

\definecolor{cvprblue}{rgb}{0.21,0.49,0.74}
\usepackage[pagebackref,breaklinks,colorlinks,citecolor=cvprblue]{hyperref}

\def\paperID{4} % *** Enter the Paper ID here
\def\confName{CVPR}
\def\confYear{2024}

\title{DELTA: Decoupling Long-Tailed Online Continual Learning}

\author{
Siddeshwar Raghavan\\
{\tt\small raghav12@purdue.edu}
\and
Jiangpeng He\\
{\tt\small he416@purdue.edu}
\and
Fengqing Zhu \\
{\tt\small zhu0@purdue.edu}
\\ % This line break will separate the authors from the institution
School of Electrical and Computer Engineering, Purdue University, West Lafayette, Indiana USA
}

\begin{document}
\maketitle
%%%%%%%%%%%%%
% Abstract %
%%%%%%%%%%%%
\begin{abstract}
A significant challenge in achieving ubiquitous Artificial Intelligence is the limited ability of models to rapidly learn new information in real-world scenarios where data follows long-tailed distributions, all while avoiding forgetting previously acquired knowledge. In this work, we study the under-explored problem of Long-Tailed Online Continual Learning (LTOCL), which aims to learn new tasks from sequentially arriving class-imbalanced data streams. Each data is observed only once for training without knowing the task data distribution. 
We present DELTA, a decoupled learning approach designed to enhance learning representations and address the substantial imbalance in LTOCL. We enhance the learning process by adapting supervised contrastive learning to attract similar samples and repel dissimilar (out-of-class) samples. Further, by balancing gradients during training using an equalization loss, DELTA significantly enhances learning outcomes and successfully mitigates catastrophic forgetting. Through extensive evaluation, we demonstrate that DELTA improves the capacity for incremental learning, surpassing existing OCL methods. Our results suggest considerable promise for applying OCL in real-world applications. Code is available online - \href{https://gitlab.com/viper-purdue/delta}{https://gitlab.com/viper-purdue/delta}.
\end{abstract}
%%%%%%%%%%%%%%%%%
% Introduction %
%%%%%%%%%%%%%%%%
\section{Introduction}
\label{sec:intro}

\begin{figure}[ht]
\begin{center}
  \includegraphics[width=1.0\columnwidth]{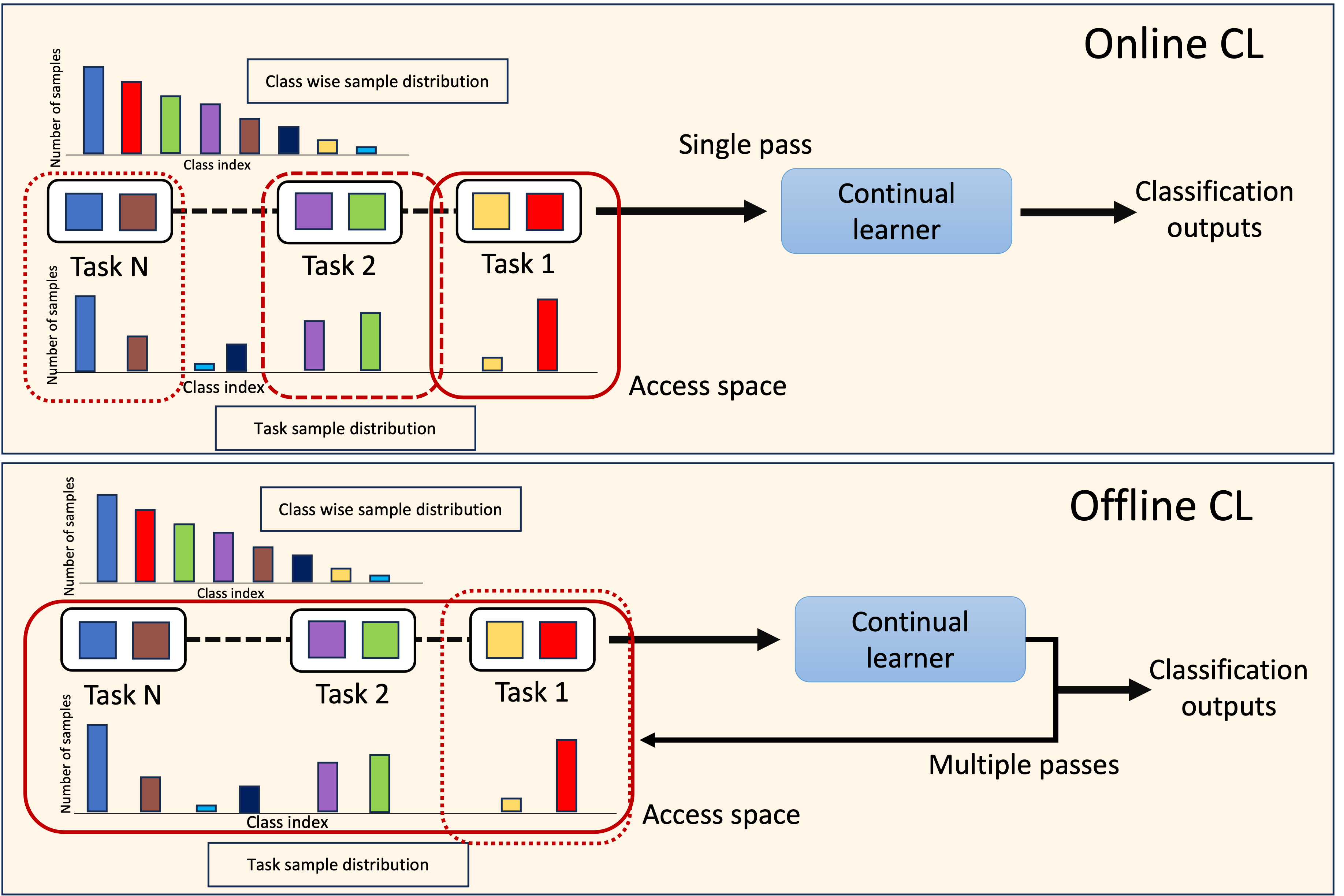}
  \caption{Illustration depicts online and offline setups for continual learning with a long-tailed distribution. In the continual learning process, tasks appear sequentially, one at a time. In the ``online" scenario, the model only accesses the current task and its distribution, while the ``offline" scenario grants access to the complete task set and their distributions. Additionally, the "online" approach involves training task data with a single pass, while the "offline" approach involves multiple passes across the entire dataset.}
  \label{fig:intro}
\end{center}
\end{figure}

The process through which humans and Artificial Intelligence (AI) systems acquire knowledge and experiences differs significantly. Over their lives, humans learn and accumulate knowledge from encountering sequential streams of temporally correlated information, mostly made up of unlabeled observations, and rarely experiencing the same scenario multiple times~\cite{lifelong_learning, CBRS}. Furthermore, humans are adept at learning, remembering knowledge, and solving multiple tasks concurrently by applying the learned knowledge. In contrast, AI systems have a smaller task focus~\cite{Krizhevsky} and a multi-stage learning approach, focusing on learning from static (non-changing) datasets through batches. Online continual learning~\cite{GSS, SCR, DVC, PRS, CBRS, PCR} strives to push the boundaries of AI by empowering agents to acquire knowledge continuously from a never-ending stream of data. However, a significant challenge continual learning systems face is how to mitigate the effects of catastrophic forgetting~\cite{LWF, EWC, catastrophic_forget_1, catastrophic_forget_2} of previously learned information.
Furthermore, recent work focuses on continual learning in the online scenario. In the online configuration~\cite{JHe_OCIL}, each data sample is used only once to train the model. Despite being more challenging, the online setting aligns with real-world limitations concerning data accessibility and computational capabilities, rendering it more appropriate for real-world applications ~\cite{He_2021_ICCVW, open_CIL_food}.

While existing online continual learning techniques have achieved impressive advancements in the context of image classification tasks, they operate under the assumption that the distribution of samples for each class is uniform~\cite{GSS, SCR, DVC, PCR, MIR, icarl}. This assumption limits their applicability in real-world scenarios where data distribution tends to be long-tailed with extreme imbalances. A concrete example of this can be observed in real-world applications such as animal species recognition~\cite{species_class_LT, Zhang2021DeepLL}, medical image diagnosis~\cite{med_LT_1, med_LT_2}, and food image classification~\cite{raghavan2024online, mao2020visual, VFN-LT}. In such cases, a minority of these images are encountered more frequently than others, leading to pronounced class imbalances. We focus on LTOCL as shown in figure~\ref{fig:intro}, wherein data sampled from such distributions emerges sequentially in a stream over time.

Though previous studies have highlighted imbalanced data distributions within the context of continual learning~\cite{PRS, CBRS, hayes_embedded}, the majority of these investigations have primarily dealt with milder imbalance ratios or investigated the offline scenario~\cite{LT-CIL}. Class imbalance signifies a scenario where the number of instances across different classes significantly varies, often leading to certain classes having a notably higher number of samples than others. However, a long-tailed distribution poses a more intricate challenge, as it involves a setup where a small subset of classes contains a substantial number of samples. In contrast, numerous classes are represented by only a limited number of samples. The distinction between online and offline CL settings is characterized by training the model once on each data sample in the online scenario, as opposed to the potential for multiple passes through the data in the offline scenario. In offline setups, the model can access the complete dataset, thereby obtaining the data distribution of classes across all tasks as seen in Figure~\ref{fig:intro}. In sharp contrast, in online scenarios, data arrives in batches, restricting the model's access solely to specific batches or subsets of data, rather than the entire dataset or complete class distributions. As a result, the data distribution for each subsequent task becomes uncertain in the context of online continual learning, and online exemplar selection becomes notably more complex within a long-tailed distribution. The presence of class-imbalanced exemplars can worsen the overfitting problem and lower the model performance during online continual learning.

Our work fills the gap between online continual learning with severe long-tailed data distribution in the image classification task. We formulate scenarios with tasks containing significant data imbalances and do not provide task identifiers during training or testing. To address these challenging scenarios, we introduce the DELTA framework. The proposed method utilizes decoupled representations in a dual-stage learning strategy, integrating contrastive learning, and Equalization Loss to re-calibrate weights in the feature space, promoting an efficient learning process.

We assess the effectiveness of our approach by comparing it against current methods in the field of OCL under long-tailed distribution conditions. Our technique consistently outperforms various experimental configurations, maintaining robustness with changes in exemplar and incremental step sizes. The main contributions of our research are summarized as follows.

\begin{itemize}
 \item We present DELTA, a dual-stage training approach that combines contrastive learning with equalization loss, tailored for Long-Tail Online Continual Learning (LTOCL) situations.
\item We propose a multi-exemplar pairing strategy to demonstrate the potential for performance enhancement in LTOCL scenarios.
\item We evaluate well-established OCL methods, compare them to DELTA in the proposed long-tailed setting under various experimental setups and report the findings.
\end{itemize}

%%%%%%%%%%%%%%%%%
% Related works %
%%%%%%%%%%%%%%%%
\section{Related Work}
\label{sec: rel_work}
In this section, we review and summarize the existing methods that are most relevant to our work, including (1) online continual learning, (2) long-tailed classification and (3) Contrastive Learning, which are illustrated in Section~\ref{subsec: ocil}, Section~\ref{subsec: long-tail review} and Section~\ref{subsec:contrastive_learning} respectively. 

\subsection{Online Continual Learning}
\label{subsec: ocil}

Continual Learning (CL) is a machine learning strategy where a model is trained to progressively incorporate new classes/ categories over time without forgetting the learned knowledge. CL has been studied under different scenarios, including (i) \textit{Online (OCL)} and \textit{Offline} settings. The former involves the model accessing solely the specific batch of data within a task~\footnote{Sequential stream of data is encountered as a set of tasks}, enabling only a single pass over the data. In contrast, the latter scenario allows the model to access the complete dataset (task-aware), permitting multiple passes over its contents~\cite{LT-CIL}. (ii) \textit{task-incremental} and \textit{class-incremental} approaches characterize another distinction, where the former necessitates a task index during both training and inference, while the latter does not. In this work, we focus on class-incremental learning in the online scenario. The objective is to learn new classes from the sequentially available data streams by using each data only once to update the model and classify all classes seen so far during the inference phase. 

Online class-incremental learning methods can be grouped into two main categories including \textit{Regularization} and \textit{Memory} based approaches. The \textbf{Regularization} methods aim to limit parameter changes to preserve learned knowledge~\cite{EWC, he2022exemplar, AGEM, GEM, JHe_OCIL}. Conversely,  \textbf{Memory} methods~\cite{SCR, DVC, PCR, GSS, PRS, CBRS, He_2022_WACV} combat catastrophic forgetting by storing task data as exemplars and replaying knowledge during learning. In the online setting, each new batch combines current task data with exemplars from memory for model updates. Memory-based methods are generally more effective than regularization-based ones ~\cite{OCIL_survey}. Furthermore, the procedures for buffer retrieval and storage differ notably between the ``online" and ``offline" settings, owing to the distinct data access constraints inherent in each setup.

Recent studies have emphasized Class Incremental Learning (CIL) setups that simulate realistic scenarios characterized by ambiguous task boundaries and imbalanced datasets~\cite{2024gradient, LT-CIL, hayes_embedded, hayes_remind}. Similarly, research has been into realistic Class Incremental Learning involving data imbalances~\cite{CBRS, PRS}. However, these investigations have predominantly centered on the offline setting~\cite{2024gradient, LT-CIL}, or often featuring moderate imbalances in the online context. Recent works have begun to explore logit adjustment within the online setting~\cite{huang2023online, CBA}. However, these investigations primarily concentrate on the conventional setting(equi-sample distribution across classes), addressing distribution shifts present in the online setting rather than tackling external, realistic data imbalances. Notably, exploring severe data imbalances within the online setting remains relatively under-explored.

\subsection{Long-Tailed Classification}
\label{subsec: long-tail review}
The long-tailed classification addresses the extreme class imbalance issue where many training classes contain a few training samples while the testing samples are class-balanced. The major challenge is the classification bias towards instance-rich (head) classes and poor generalization ability in classifying instance-rare (tail) classes. As the long-tailed classification has been widely studied over decades~\cite{zhang2021deep}, we review the existing (\textit{i.e. }end-to-end) approaches that are most related to our work. \textbf{Re-sampling} based methods aim to address class-imbalance issues by generating balanced training distribution. The typical work includes over-sampling~\cite{ROS} the instance-rare classes and under-sampling~\cite{RUS_ROS, he_2023icip} the instance-rich classes. The most recent work~\cite{CMO, VFN-LT, he2023long} further applies CutMix~\cite{yun2019cutmix} as data augmentation to mitigate the over-fitting and under-fitting issues caused by naive sampling strategies. \textbf{Re-weighting} based method balances the loss gradients by assigning higher weights on instance-rare and lower weights on instance-rich classes or data samples. Specifically, the class level weights in~\cite{inverse_frequency_1, inverse_frequency_2} are generated based on the inverse of class frequency. Furthermore, there are Class-balanced loss~\cite{class_bal_loss}, label-distribution-aware-margin loss~\cite{LDAM}, balanced softmax~\cite{BSLoss} and LADE loss~\cite{LADE}, which aim to balance the loss gradients. \textbf{Two-stage methods}~\cite{decoupling1, devil_decouple,bilateral_decouple, decouple_2} focus on decoupling imbalanced feature learning from balanced classifier learning~\cite{decouple_3}. However, all the existing methods require knowing the data distribution in the entire training set, which is not feasible in the online continual learning scenario where the new data comes sequentially in a stream over time. In this study, we draw inspiration from re-sampling and re-weighting approaches used in Long-Tail image classification to design an Equalization loss to mitigate the severe imbalances present without knowing the entire distribution of the dataset.

\subsection{Contrastive Learning}
\label{subsec:contrastive_learning}
Contrastive learning effectively learns meaningful data representations by pulling similar samples closer and pushing dissimilar ones apart. This approach enables the model to uncover and leverage the underlying structures and semantics of the data, resulting in representations beneficial for various downstream tasks~\cite{contrastive_survey, khosla2020supervised, chen2020simple,Rai_2021_CVPR, contrastive_what_nt}. Researchers have demonstrated that contrastive learning is applicable in both supervised and unsupervised settings. In unsupervised contrastive learning, data augmentations act as similar samples, and randomly selected samples from the target batch act as dissimilar samples~\cite{pmlr-contrastive, unsup-contrastive}. Conversely, in supervised contrastive learning, samples from the same class are treated as similar, and those from different classes are considered dissimilar~\cite{khosla2020supervised, pmlrsuper-contrastive}.

%%%%%%%%%%%%%%%%%
% Prob setup   %
%%%%%%%%%%%%%%%%
\section{Preliminaries for Long-Tailed Online Continual Learning}
\label{subsec:prelim}

We examine a data distribution characterized by a significant long-tailed nature in the context of online continual learning for supervised image classification tasks. The long-tailed distribution follows an exponential decay in sample sizes across classes~\cite{LDAM, BSLoss}. This decay is parameterized by $\rho$, the ratio between the most and least appearing classes. In the OCL setting, the model encounters a continuous stream of data at each task $t$. In this setup, at any task $t$, $k^t$ represents the number of classes, and $n_j^t$ represents the number of samples in class $j$. Thus, the number of classes learned up to task $t$ can be represented as $k^{1:t}$. The training samples at task $t$, denoted as ${\mathcal{X}_t} = \left \{ x_i^t, y_i^t \right \}; i \in \left \{1,2,..,k^{t} \right \} $, are non-independent and identically distributed (non i.i.d), drawn from the current distribution $D_t$ (${x}_i^t$ represents an image and ${y}_i^t$ represents its corresponding label). Within the Online Continual Learning (OCL) setting, the model's access is confined to the data of a specific batch within task $t$, given the sequential flow of data. Consequently, it remains aware solely of the data distribution up to that point. This distribution, referred to as $D_t$, is subject to change between tasks, transitioning from $D_t$ to $D_{t+1}$. 

ER methods involve the utilization of a fixed-size memory buffer denoted as $B$. This buffer stores a limited set of samples from the learned tasks. This stored subset is drawn upon for knowledge rehearsal during the learning of subsequent tasks. After training on each task $t$ in the continual learning process, the model is tested on a balanced held-out test set denoted as $\overline{\mathcal{X}_t} = \left \{ \overline{x}_i^t, \overline{y}_i^t \right \}; \, i \in \left \{1,2,..,k^{t} \right \}$. This test set includes only the classes encountered so far. The total number of training samples in task $t$ is denoted as $n^t$, and it satisfies the condition $\sum_{j=1}^{k^t} n_{j}^{t} = n^t$.

%%%%%%%%%%%%%%%%%
% Methodology  %
%%%%%%%%%%%%%%%%

\section{Method}
\label{sec:method}
Decoupling representation learning from the classification task has been shown effective in two-stage network architectures tackling the problem of long-tailed recognition~\cite{beierxERM, LT_2stage, LT_2stage_tang}. This concept is adapted for the OCL setting with long-tailed data by developing a two-stage approach that incorporates contrastive learning in the first stage and implements an Equalization Loss in the second stage to address the pronounced imbalances in the data.

The structure of DELTA is depicted in figure~\ref{fig:delta_method}. We have included the pseudo code in the supplementary material.
\begin{itemize}
    \item \textbf{Stage 1} comprises contrastive learning in the supervised setting due to the availability of labels in OCL. The motivation to utilize contrastive learning in the long-tailed setting is to cluster similar samples and push apart dissimilar samples to aid in effective feature learning.
    \item \textbf{Stage 2} consists of training just the classification layer of the network using an Equalization Loss to re-weight the samples, to fine-tune and obtain a better classifier. 
\end{itemize}

\begin{figure*}
    \centering
    \includegraphics[width=2\columnwidth]{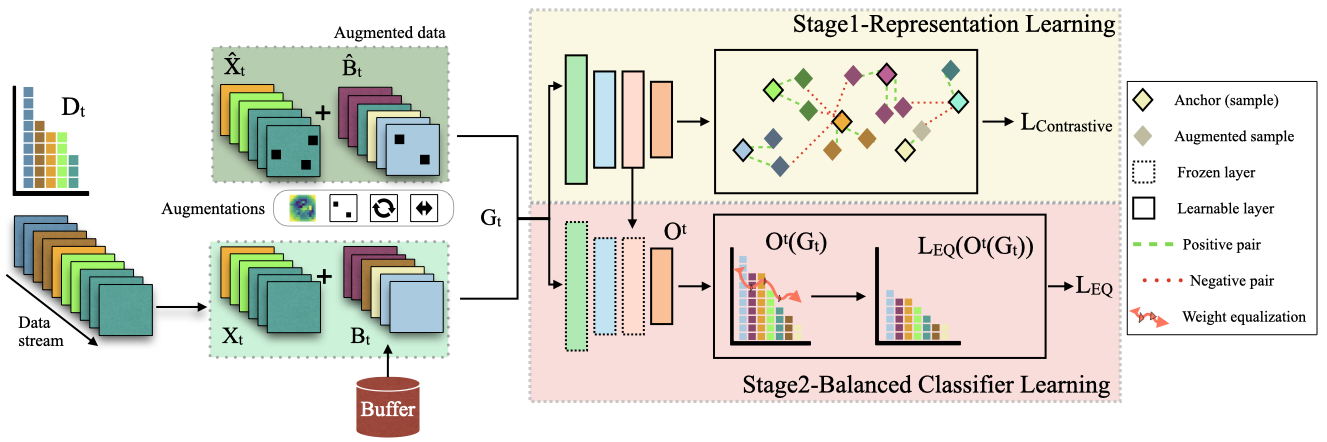}
    \caption{ An overview of the DELTA framework: At task $t$, the current batch of samples($X_t$) and samples retrieved from the memory buffer ($B_t$) undergo augmentation ($\hat{X}_t$, $\hat{B}_t$) and are then combined ($G_t$). This combined data is directed sequentially through a dual-stage training pipeline. In the first stage, the framework utilizes contrastive learning to generate effective data representations involving a contrastive loss (\(L_{contrastive}\)). During the second stage, the learning approach is decoupled by keeping all layers frozen except for the classification layer (\(O^t\)). This targeted training employs the weight equalization loss (\(L_{EQ}\)) to train a balanced classifier and reduce the shift in future data representations.}
    \label{fig:delta_method}
\end{figure*}

\subsection{Stage 1 - Representation Learning}
\label{subsec:stage1}

In Online Continual Learning (OCL), where data undergoes single-pass processing and exhibits a pronounced imbalance, particularly in long-tail distributions, effectively learning the underlying feature representations is essential.

Various methodologies within contrastive learning, such as Barlow Twins~\cite{zbontar2021barlow}, SimSiam~\cite{chen2020simsiam} and BYOL~\cite{BYOL}, primarily focus on leveraging positive samples for learning. However, our methodology draws inspiration from SimCLR~\cite{chen2020simple}, which employs both positive and negative samples advantageously. In long-tailed distributions, where many classes have few samples, the diversity of negative samples from the more populated classes can help the model learn more discriminating features for the underrepresented classes. Additionally, SimCLR heavily relies on aggressive data augmentation strategies to generate positive pairs. This approach can help mitigate the effects of class imbalance by ensuring that the model learns robust features for each class, regardless of its frequency in the dataset. There are three main components in the contrastive learning pipeline~\cite{SCR, chen2020simple}:
\begin{itemize}
    \item Data augmentation on the input sample is performed, denoted as $\widehat{x} = Aug(x)$.
    \item The encoder network, represented as $Encoder(.)$, transforms an image sample into a vector embedding, $e = Encoder(x) \in \mathbb{R}^{D_N}$, with normalization to the unit space in $\mathbb{R}^{D_N}$.
    \item The projection network, indicated by $Projection(.)$, takes the embedding and maps it to a projected vector, $v = Projection(e) \in \mathbb{R}^{D_P}$, which is then normalized using the L2 norm.
\end{itemize}
The contrastive loss is defined as,

\begin{align}
    \scalebox{0.85}{\(L_{contrastive}(Z_T)\)} = \sum_{j \in T}\frac{-1}{|P(j)|}\sum_{p \in P(j)}\frac{exp(v_j \cdot v_p/\tau)}{\sum_{k \in A(j)}exp(v_j \cdot v_p/\tau)}
\end{align}

$B_T = B_t \cup Aug(B_t)$ represents the samples obtained from the buffer for task $t$, where $B_t$ is the set of original buffer samples and $Aug(B_t)$ is their augmented versions. The index set $T$ identifies elements of $B_T$ excluding the $i$th sample. $\mathcal{X}_t(i)$ refers to all elements in $B_T$ excluding sample $i$. The function $P(j)$ identifies the set of positive samples in $B_T$ that share the same label as sample $i$, but does not include sample $i$ itself. The parameter $\tau \in \mathbb{R}$ serves as a temperature factor to modulate class separation, and the symbol $\cdot$ denotes the dot product operation.

\subsection{Stage 2 - Balanced Classifier Learning}
\label{subsec:EQ loss}
In the DELTA framework's second stage, the goal is to establish a balanced training environment for the classifier by decoupling the learning of feature representations from the classification task. 
In scenarios with long-tailed distributions, where class distribution is highly imbalanced, a model trained to minimize empirical risk often underperforms on a balanced test dataset due to distribution mismatch. Cross-entropy loss in such scenarios can result in learning-biased batches. While many Online Class Incremental Learning (OCL) methods~\cite{OCIL_survey}, employ Cross-Entropy (CE) loss, this loss function inherently favors the classification of majority classes, potentially at the expense of minority class accuracy. 
To counteract the issue of biased batch learning, we propose the Equalization Loss ($L_{EQ}$), an innovative technique designed for online continual learning environments. This method is inspired by the balanced softmax loss~\cite{BSLoss}, which requires the entire data distribution of the dataset. In Long-Tailed Online Continual Learning (LTOCL) settings, we cannot access the entire dataset but obtain the data via continuous streams.  Thus, we incorporate a task-specific distribution vector, \(P(k^t)\), which is updated after encountering a data stream within a task, as shown in Equation~\ref{eqn:temp_dist_vec}.  In this context, \(D_t\) represents the distribution of samples for task \(t\), encompassing both the training inputs and the exemplars retrieved from the buffer.
\begin{align}
    \label{eqn:temp_dist_vec}
    D_{t} &= [n_1, n_2, .... n_h] \\
    \textrm{$P(k^{t})$} &= [\frac{n_1}{\sum_{i=1}^{h}n_i}, \frac{n_2}{\sum_{i=1}^{h}n_i},....,\frac{n_h}{\sum_{i=1}^{h}n_i}]
\end{align}
where where $\{n_h; h=1,2,..,k^{1:t} \}$ denotes the number of samples in class $h$.
This vector dynamically characterizes the sample distribution within each incoming training batch for a given task $t$. 

The logits at the output stage of the classifier is expressed as $O^t(I_x) = [O^1(I_x), O^2(I_x),...,O^{1:t-1}(I_x),...,O^t(I_x)]$, respectively, where $I_x$ represents the input image sample to the classifier. For each incoming training batch, we calculate the temporary probability distribution. We then adjust the gradients to reflect this distribution by incorporating $P(k^t)$ as a prior vector in the output, a process detailed in Equation~\ref{eqn:balsoft}.
\vspace{-0.2cm}
\begin{align}
\label{eqn:balsoft}
\scalebox{0.85}{\(\mathcal{L}_{EQ}(O^t(I_x)) = \sum_{i=1}^{k^{1:t}} - I_{y_{(i)}}\sigma([\log[(P(k^t)]+O^t(I_x))])\)}
\end{align}

where $I_y$ is the corresponding label for input $Ix$ and $\sigma$ represents the softmax function. Therefore, the more frequent class of the current training batch with a larger value in the prior vector $P(k^t)$ achieves smaller gradients when we compute the cross-entropy using the adjusted logits and vice versa. Therefore, effectively address the biased loss gradients issue in the long-tailed online continual learning without requiring knowing the data distribution beforehand. 

The cornerstone of our Equalization Loss ($L_{EQ}$) is to correct gradient disparities by utilizing the temporary distribution vector, thereby bridging the gap between the training data's long-tailed nature and the testing data's balanced nature. In turn, it improves the model's capacity to effectively retain and generalize knowledge across different classes.

The overall loss in stage1 is represented in Equation \ref{eqn:total_loss_stage1}.
\vspace{-0.2cm}
\begin{align}
\label{eqn:total_loss_stage1}
\scalebox{0.95}{\(\mathcal{L}_{stage1}(O^t(G_t)) = \mathcal{L}_{contrastive}(O^t(G_t))\)} 
\end{align}
$G$ represents the input samples, the buffer retrieved samples and their augmented versions. $G_t = B_t \bigcup \widehat{B_t} \bigcup \mathcal{X}_t \bigcup \widehat{\mathcal{X}_t}$
In \textbf{stage2}, as shown in figure~\ref{fig:delta_method}, we freeze all the layers except the classification layer to train the network using the Equalization Loss to obtain the best classification accuracy, represented as,
\vspace{-0.2cm}
\begin{align}
\label{eqn:total_loss_stage2}
\scalebox{0.95}{\(\mathcal{L}_{stage2}(O^t(G_t)) = \mathcal{L}_{EQ}(O^t(G_t))\)}
\end{align}

\subsection{Multi-Exemplar Learning}
\label{subsec:multiexemplars}
The discrepancies between long-tailed training distributions and balanced test distributions pose a significant challenge, leading to a bias in the learning algorithm towards the training data due to uneven sample sizes exacerbated in the online setting. To address this, we propose an exemplar selection strategy pairing more than one exemplar with each training sample to balance the batch composition and mitigate bias. This approach preserves the data's inherent randomness and ensures that exemplar representation aligns with the distribution vector. As shown in figure~\ref{fig:abl_exemplars}, the x-axis denotes the number of exemplars paired per input data sample. 

Our work is among the earliest in exploring multi-exemplar pairing within the context of OCL. Traditional approaches~\cite{GSS, ASER, MIR, onpro, CBRS, PRS, LT-CIL} typically limit themselves to matching a single exemplar from the memory buffer with each sample from the current batch. However, this may not be advantageous when the number of tasks increases or data exhibits a high variability. This practice often results in sub-optimal performance due to the one-off nature of the process. 
%Methods, like Repeated Augmented Rehearsal~\cite{RAR} and GSS~\cite{GSS}, suggest batch-level multi-iterations that improve performance while increasing the compute overhead. 
Moving beyond the one-exemplar-per-sample restriction for each new batch opens a new direction in OCL. Our DELTA method incorporates data augmentations before the training phase, which helps mitigate overfitting despite multiple encounters with previously learned samples from the buffer. Moreover, this repeated exposure to past samples plays a crucial role in combating catastrophic forgetting and enhances the accuracy of learned representations.

The empirical risk for any given task $t$ can be formulated as
\begin{align}
    \scalebox{1.2}{\(R_{emp}f(\theta) = \mathbb{E}_{(G_{t\_x}, G_{t\_y})\sim D_{{G}_t}}\mathcal{L}(f(G_{t\_x}; \theta), G_{t\_y})\)}
\end{align}
Where $R_{emp}f(\theta)$ signifies the empirical risk of the model $f$ for a given task $t$, $\mathbb{E}$ represents the expected value, $\mathcal{L}$ denotes the loss function, and $G_{t\_x}, G_{t\_y}$ correspond to the images and their respective labels, consisting of input, buffer data and their augmented pairs. However, the empirical risk evaluated on the training dataset does not equate to the true risk on the test dataset, given that $D_{\mathcal{X}} \neq D_{\mathcal{\overline{X}}}$, where $D_{\mathcal{X}}$ and $D_{\mathcal{\overline{X}}}$ denote the distributions of the training and test sets, respectively.
\begin{align}
    \scalebox{1.2}{\(R_{true}f(\theta) = \mathbb{E}_{(\overline{x}^t, \overline{y}^t)\sim D_{\mathcal{\overline{X}}}}\mathcal{L}(f(\overline{x}^t; \theta), \overline{y}^t)\)}
\end{align}

Increasing the number of paired exemplars and incorporating augmentations while balancing each batch are aimed at enhancing the model's robustness to input variations and improving its generalization to a held-out test set. By training with a dataset enriched to reflect the data's inherent variability better, the continual learner is less biased toward the long-tailed nature of the training data. This approach leads to more accurate and unbiased estimates of the gradient and reduces the variance in model updates, facilitating smoother convergence towards the empirical loss minimum. We aim to balance the data distribution within each batch more accurately; the likelihood of the model overfitting to specific training data features is decreased, resulting in improved generalization performance on unseen data.

%%%%%%%%%%%%%%%%%
% Experiments  %
%%%%%%%%%%%%%%%%

\section{Experiments}
\label{sec:exp}

% ACCURACY TABLE
\begin{table*}[]
\centering
\resizebox{1.0\textwidth}{!}{%
\begin{tabular}{l|cccccc||cccccc}
\hline
Methods & \multicolumn{6}{c||}{CIFAR100-LT} & \multicolumn{6}{c}{VFN-LT} \\ \hline
 & 20 tasks & 20 tasks & \multicolumn{1}{c|}{20 tasks} & 10 tasks & 10 tasks & 10 tasks & 15 tasks & 15 tasks & \multicolumn{1}{c|}{15 tasks} & 7 tasks & 7 tasks & 7 tasks \\ \hline
 & M=0.5K & M=1K & \multicolumn{1}{c|}{M=2K} & M=0.5K & M=1K & M=2K & M=0.5K & M=1K & \multicolumn{1}{c|}{M=2K} & M=0.5K & M=1K & M=2K \\ \hline
OnPRO[ICCV '23] & 14.02 $0.44$  &16.28 $\pm$ 0.81  & \multicolumn{1}{c|}{18.01 $\pm$ 0.22} & 16.53 $\pm$ 0.55  & 16.92 $\pm$ 0.08  &18.85 $\pm$ 0.32  & 11.93 $\pm$ 0.04  & 12.77 $\pm$ 0.07  & \multicolumn{1}{c|}{13.50 $\pm$ 0.05} &\textbf{8.02 $\pm$ 0.60}  & 9.38 $\pm$ 0.21  & 11.84 $\pm$ 0.49  \\
SCR[CVPRW '21] & 12.22 $\pm$ 0.72 & 13.48 $\pm$ 0.90 & \multicolumn{1}{c|}{15.88 $\pm$ 0.79} & 16.65 $\pm$ 0.90 & 17.02 $\pm$ 0.77 & 17.58 $\pm$ 0.66 & 11.55 $\pm$ 0.17 & 11.82 $\pm$ 0.10 & \multicolumn{1}{c|}{12.39 $\pm$ 0.73} & 7.71 $\pm$  0.49 & 9.19 $\pm$  0.46 & 9.48 $\pm$  0.47 \\
ASER[AAAI '21] & 8.86 $\pm$ 0.30 & 7.86 $\pm$ 0.61 & \multicolumn{1}{c|}{8.18 $\pm$ 0.31} & 12.68 $\pm$ 0.70 & 13.76 $\pm$ 0.01 & 15.90 $\pm$ 0.91 & 6.85 $\pm$  0.34 & 7.61 $\pm$  0.38 & \multicolumn{1}{c|}{7.22 $\pm$  0.36} & 7.46 $\pm$ 1.18 & 7.52 $\pm$ 1.09 & 6.35 $\pm$ 0.19 \\
PRS[ECCV '20] & 7.61 $\pm$ 0.09 & 7.54 $\pm$ 0.21 & \multicolumn{1}{c|}{7.03 $\pm$ 0.13} & 7.34 $\pm$ 0.92 & 8.95 $\pm$ 0.33 & 9.01 $\pm$ 0.39 & 7.17 $\pm$ 0.83 & 8.72 $\pm$ 0.15 & \multicolumn{1}{c|}{8.39 $\pm$ 0.19} & 7.85 $\pm$ 0.50 & 8.66 $\pm$ 0.22 & 9.21 $\pm$ 0.30 \\
CBRS[ICML '20] & 8.51 $\pm$ 0.19 & 8.66 $\pm$ 0.61 & \multicolumn{1}{c|}{8.91 $\pm$ 0.33} & 9.50 $\pm$ 0.48 & 7.22 $\pm$ 0.43 & 7.31 $\pm$ 0.08 & 8.12 $\pm$ 0.94 & 8.35 $\pm$ 0.33 & \multicolumn{1}{c|}{8.18 $\pm$ 0.44} & 7.52 $\pm$ 0.11 & 7.64 $\pm$ 0.08 & 7.92 $\pm$ 0.34 \\
%DER++[NeurIPS '20] &  &  & \multicolumn{1}{c|}{} &  &  &  &  &  & \multicolumn{1}{c|}{} &  &  &  \\ 
GSS[NeurIPS '19] & 5.16 $\pm$ 0.10 & 5.22 $\pm$ 0.22 & \multicolumn{1}{c|}{5.09 $\pm$ 0.21} & 8.97 $\pm$ 0.65 & 10.12 $\pm$ 0.02 & 9.96 $\pm$ 0.47 & 5.86 $\pm$ 0.30 & 6.01 $\pm$ 0.91 & \multicolumn{1}{c|}{5.86 $\pm$ 0.06} & 5.92 $\pm$ 0.54 & 4.30 $\pm$ 0.22 & 4.66 $\pm$ 0.60 \\ 
LT-CIL(offline) & 3.01 $\pm$ 0.77 & 2.67 $\pm$ 0.04 & \multicolumn{1}{c|}{2.43 $\pm$ 0.02} & 1.76 $\pm$ 0.11 & 2.36 $\pm$ 0.25 & 3.76 $\pm$ 0.22 & 1.82 $\pm$ 0.45 & 2.02 $\pm$ 0.44 & \multicolumn{1}{c|}{2.38 $\pm$ 0.08} & 3.08 $\pm$ 0.71 & 2.92 $\pm$ 0.04 & 1.99 $\pm$ 0.31 \\ \hline
DELTA (ours) & \textbf{16.53 $\pm$ 0.01} & \textbf{17.71 $\pm$ 0.11} & \multicolumn{1}{c|}{\textbf{19.93 $\pm$ 0.07}} & \textbf{20.25 $\pm$ 0.71} & \textbf{21.06 $\pm$ 0.23} & \textbf{22.47 $\pm$ 0.51} & \textbf{12.5 $\pm$ 0.01} & \textbf{13.45 $\pm$ 0.02} & \multicolumn{1}{c|}{\textbf{13.84 $\pm$ 0.01}} & 8.00 $\pm$ 0.39 & \textbf{10.41 $\pm$  0.52} & \textbf{12.84 $\pm$  0.54} \\ \hline
\end{tabular}
}
\caption{\textbf{Average Accuracy ($\%$) $\pm$ standard deviation} ($\uparrow$) in the \textbf{long tailed} scenario on Split CIFAR-100-LT, VFN-LT with single exemplar pairing. The best accuracy results are highlighted in \textbf{boldface}}
\label{tab:lt_acc}
\end{table*}

%\begin{figure}
%    \centering
%    \includegraphics[width=0.8\columnwidth]{CVPRW4/images/10tasks.png}
%    \caption{Task wise accuracy of compared methods in LTOCL on CIFAR100-LT with 10 tasks, 2K memory and $\rho =0.01$ }
%    \label{fig:10tasks}
%\end{figure}

In this section, we first introduce the datasets and our experimental setup. Then, we comprehensively analyze the performance of the current OCL methods in the conventional and the LT setup. Finally, we conduct ablation studies to show the effectiveness of each component in our proposed framework. 

\subsection{Datasets}
We use two publicly available datasets, CIFAR-100~\cite{Cifar100} (100 classes), and the VFN-LT~\cite{VFN-LT} (74 classes). We create the long-tailed version of Split CIFAR-100 with the imbalance factor $\rho = 0.01$, where $\rho$ represents the ratio between most frequent and least frequent classes~\cite{LT-CIL}. The smaller the value of $\rho$, the more pronounced the imbalance. Overall, the Split CIFAR-100 has over 10K training images with a maximum of 500 images and a minimum of 5 images per class. The VFN-LT dataset reflects the real-world food distribution compared to other datasets. It is long-tailed, containing over 15,000 training images across 74 classes, representing commonly consumed food categories in the United States based on the WWEIA database~\footnote{https://data.nal.usda.gov/dataset/what-we-eat-america-wweia-database}.
\subsection{Implementation Detail}
% describe model, training parameters, batch size, ... etc. also the evaluation metrics, exemplar size. 

Our implementation is PyTorch~\cite{pytorch} based. We use ResNet-32 for Split CIFAR-100 and ResNet-18 for VFN-LT dataset, which acts as the Encoder network and for the projection network we use a fully connected layer to map the representations from the encoder to 128-dimensional latent space~\cite{chen2020simple}. We train the networks from scratch and split the datasets using fixed seed 1993. The input image size set for Split CIFAR-100 is $32 \times 32$, and $224 \times 224$ for VFN-LT following the settings suggested in \cite{OCIL_survey}. For CIFAR-100-LT, we evaluate two configurations: one with 20 tasks, each comprising five unique classes, and another with 10 tasks, each containing ten unique classes. For VFN-LT, the division is into 15 tasks, where the initial task contains four classes, and each subsequent task consists of five classes, and another configuration with seven tasks, where the first task involves 14 classes, and each of the following tasks includes ten classes. We use a stochastic gradient descent optimizer with a fixed learning rate of 0.1 and a weight decay of $10^{-4}$. The training batch size is 16, and the testing batch size is 128. The data (except exemplars) is seen only once by the model to train for all the experiments. For our experiments, we implement three memory buffer sizes (0.5K, 1K, and 2K) for various experience replay methods and configure single exemplar pairing. For the buffer, we use random sampling and reservoir update mechanism extended from~\cite{OCIL_survey}. We set the temperature parameter $\tau$ as 0.09, obtained via grid search. We run each experiment 5 times and report the average accuracy in Table~\ref{tab:lt_acc}. We run all our experiments on a single NVIDIA A40 GPU. 

In our work, we employ the publicly accessible implementations of all existing OCL methods, as referenced in~\cite{OCIL_survey, onpro, PRS, CBRS, LT-CIL}, for our comparative analysis. We have considered an offline OCL method~\cite{LT-CIL} for comparison and we implement it in the online setting by running the method for one epoch and consider only the task-agnostic setting for fair evaluation. Additionally, we have integrated a long-tailed data loader into this framework and have implemented our proposed method, DELTA, for a comprehensive comparison.

\subsection{Evaluation Metrics}
In this work, we employ Average Accuracy to evaluate performance. Average Accuracy assesses the overall performance across the testing sets from previously encountered tasks. We have included the forgetting metrics in the supplementary section. 
Let $a_{i,j}$ be the model's performance on the held-out testing set of task $j$ after the model is trained from task $1$ to $i$~\cite{OCIL_survey}. For a total of $T$ tasks :
\begin{equation}
    \label{eqn:avg_acc}
    \text{Average Accuracy -} A_T = \frac{1}{T}\sum_{j=1}^{T}a_{T, j}
\end{equation}

\subsection{Discussion of Results}
\begin{figure*}[ht!]
    \centering
    \includegraphics[width=0.92\textwidth]{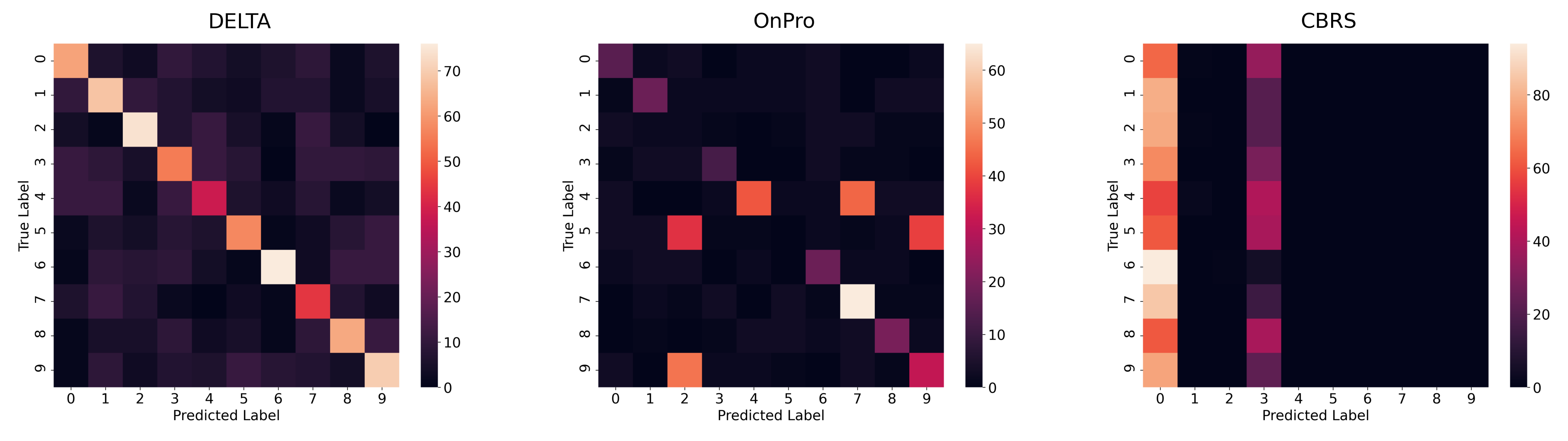}
    \caption{Confusion matrices for DELTA, OnPro~\cite{onpro}, and CBRS~\cite{CBRS} on CIFAR100-LT with a memory buffer of 2,000 show distinct patterns. Single-stage methods(OnPro, CBRS) are prone to a bias towards recent tasks, particularly with long-tailed samples, often misclassifying numerous samples as belonging to the latest task classes. DELTA exhibits a reduced bias thanks to its unique decoupled learning architecture that incorporates a contrastive learner and employs an equalization loss.}
    \label{fig:CM_methods}
\end{figure*}

In this section, we evaluate the discussed OCL methods under the long-tailed condition (with \(\rho\) = 0.01), using varying memory buffer sizes across two datasets: Split CIFAR-100 \cite{Cifar100}, and the VFN-LT \cite{VFN-LT} dataset. Table~\ref{tab:lt_acc} reveals that the average accuracy of current OCL methods is comparatively low in these long-tailed scenarios. In contrast, as shown in Table~\ref{tab:ablation_imbalance}, these methods demonstrate improved performance in conventional settings (\(\rho = 1\)) where the class distribution is balanced. The diminished performance in long-tailed scenarios is because existing approaches are not designed to handle the significant imbalances commonly present in real-world data.
Learning tail classes is challenging in the online scenario with a single pass over the data, often resulting in reduced accuracy, as evidenced in Table~\ref{tab:lt_acc}. Among the existing OCL methods, SCR and CBRS outperform others. SCR achieves this through tightly clustering related class embeddings and using an NCM classifier, while CBRS benefits from class-balanced sampling. Variations in exemplar size, ranging from 0.5K to 2K, reveal that the performance of existing OCL methods is inconsistent with increased buffer size. In contrast, our approach demonstrates consistency and resilience against variations in exemplar size and task sizes. We attribute the performance gain to our DELTA method, which incorporates a dual-stage decoupled learning pipeline with contrastive learning and equalization loss structure. Utilizing contrastive learning, we effectively cluster long-tailed samples and integrate EQ loss results in more accurately learned representations. In the subsequent stage, the decoupling of representation learning allows for a more effective classification task facilitated by the EQ loss ($L_{EQ}$) that addresses data imbalances. We showcase our method is significantly less biased towards the long-tailed data after training on the last task in Figure~\ref{fig:CM_methods}, whereas other methods are biased towards the classes appearing in the task. Our approach demonstrates consistent performance in terms of accuracy (Table~\ref{tab:lt_acc})across varying buffer sizes, task sizes, and imbalance ratios(Table~\ref{tab:ablation_imbalance}).

\subsection{Ablation Study}
\label{subsec:ablation}
\textbf{Effectiveness of dual-stage approach DELTA and Equalization Loss}
To verify the effectiveness of the dual-stage approach, we compare the performance in various imbalance ratios ranging from mild to severe imbalance representative of real-world scenarios. We showcase the consolidated results in Table~\ref{tab:ablation_imbalance}. Additionally, we replace the Equalization Loss (detailed in section~\ref{subsec:EQ loss}) with Cross Entropy (CE) to showcase the enhancements, whereas CE is biased toward classes with more number of samples as it averages the loss over all samples and do not effectively promote learning in the long-tailed scenario as shown in Table~\ref{tab:ablation_CE_EQ}. In comparison Equalization loss effectively adjusts the prediction scores based on class frequency making the model less biased towards the majority classes leading to improved performance.

\begin{table}[h]
    \centering
    \scalebox{0.80}{
    \begin{tabular}{l|ccc}
        \hline
        Imbalance ratio ($\rho$) &SCR  &CBRS  & DELTA \textit{(ours)} \\
         \hline
         0.005 & 3.59 $\pm$ 0.64 &  7.60 $\pm$ 0.06  & \textbf{18.02 $\pm$ 0.79} \\
         0.03 & 4.11 $\pm$ 0.42  &  8.88 $\pm$ 0.47 & \textbf{20.21 $\pm$ 0.22} \\
         0.07 & 8.34 $\pm$ 0.04  &  9.47 $\pm$ 0.61  & \textbf{23.60 $\pm$ 0.09}  \\
         0.1 & 6.12 $\pm$ 0.35  &  10.26$\pm$ 0.38  & \textbf{24.28 $\pm$ 0.60}  \\
         1.0 (Conventional) & 20.74 $\pm$ 0.40 & 15.79 $\pm$ 0.33  & \textbf{33.23 $\pm$ 0.97} \\
        \hline  
        \hline
    \end{tabular}
    }
    \caption{Ablation study for average accuracy (\%) with different imbalance ratios on Split CIFAR-100 for~\emph{long-tailed} distributions with a fixed exemplar size 2K and with single-exemplar pairing. Compared against the best performing methods. Smaller the value of $\rho$, greater the imbalance between most frequent and least frequent classes.}
    \label{tab:ablation_imbalance}
\end{table}

\begin{table}[h]
    \centering
    \scalebox{.80}{
    \begin{tabular}{ccc|cc}
        \hline
        $L_{contrastive}$ & $L_{CE}$ & $L_{EQ}$ &\textbf{Split CIFAR-100}&\textbf{VFN-LT} \\
         \hline
          \checkmark & \checkmark & & 16.62 $\pm$ 0.91 & 9.91 $\pm$ 0.28 \\
          \checkmark & & \checkmark & \textbf{19.93 $\pm$ 0.07} & \textbf{13.84 $\pm$ 0.01}\\
        \hline  
        \hline
    \end{tabular}
    }
    \caption{Ablation study for average accuracy (\%) on the loss function used in stage 1 and stage 2 of DELTA in addition to the Contrastive Loss ($L_{contrastive}$). }
    \label{tab:ablation_CE_EQ}
\end{table}

\begin{figure}
    \centering
    \includegraphics[width=1.0\columnwidth]{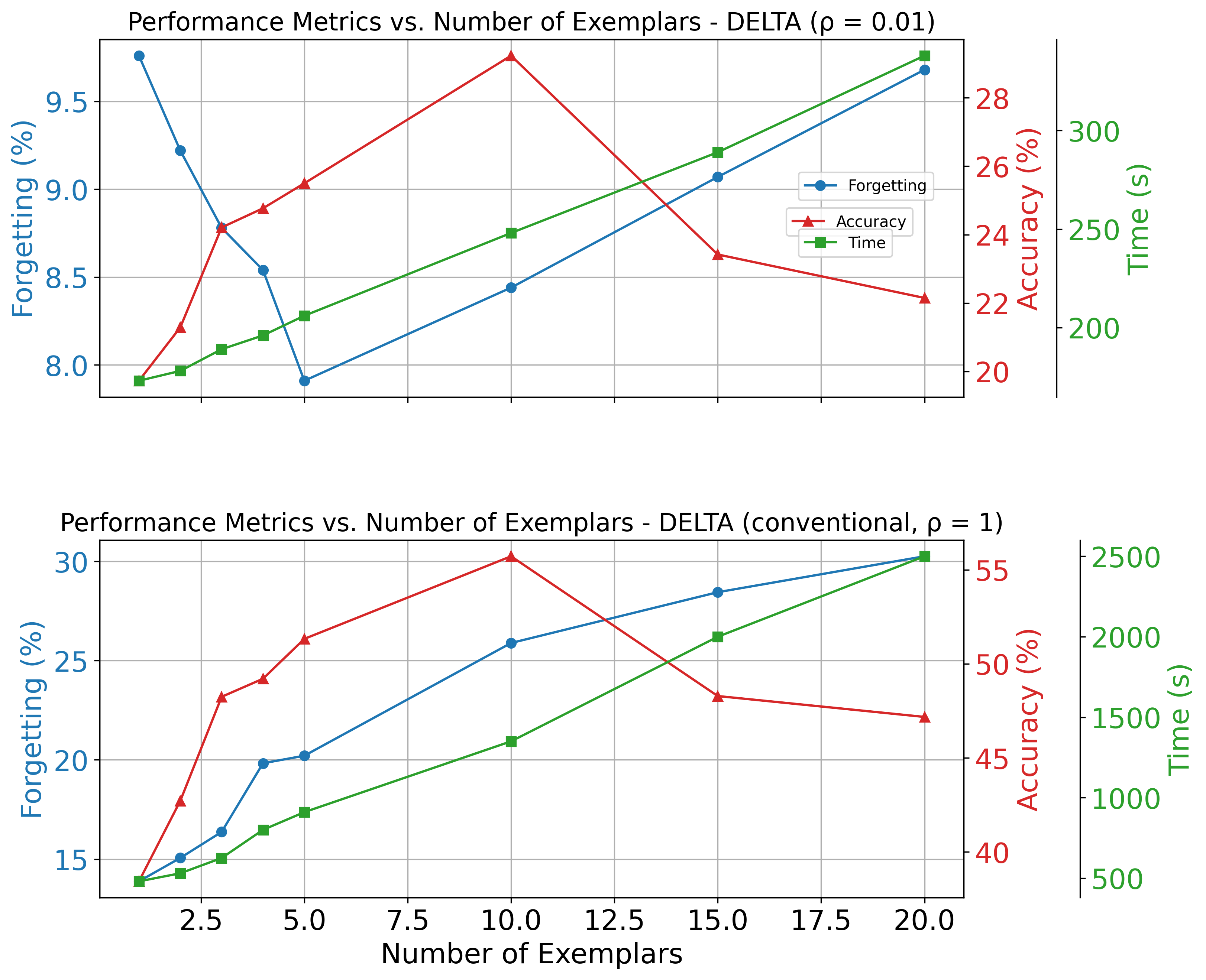}
    \caption{Performance of DELTA at $\rho = 0.01$ (top), and DELTA at $\rho = 1$ (conventional) with an increasing number of paired exemplars. The graph displays CIFAR100-LT utilizing a 2K buffer across 20 tasks.}
    \label{fig:abl_exemplars}
\end{figure}

% 4. Use of CE in stage 2 instead of balanced softmax to show effectiveness of Balanced softmax.
\textbf{Analysis of multi-exemplar pairing}
As explained in Section~\ref{subsec:multiexemplars}, we analyze the effect of multi-exemplar pairing within our DELTA method and demonstrate the superiority of our method in terms of learning accuracy, memory efficiency, and time trade-off (during training) through multi-exemplar pairing. For every input image sample, we pair it with one or more exemplars from the buffer, effectively enlarging the training batch size. This strategy aims to mitigate the variance of gradient estimates compared to smaller data batches. This reduced variance means the direction of the gradient descent steps is more consistent and stable, leading to more reliable training progress, even if the data within those batches is not perfectly balanced. The depicted figure~\ref{fig:abl_exemplars} presents the average accuracy, average forgetting, and the duration of training for 20 tasks. We observe a decline in performance when each input sample is paired with more than ten exemplars, indicating the onset of overfitting beyond this threshold. 
% Multi-exemplar pairing can be extended to other OCL methods and we've depicted using SCR in figure~\ref{fig:abl_exemplars} (middle) to showcase the improved performance achieved when paired with more exemplars in the long-tailed scenario. 
It is to be noted that, the performance is improved even in the conventional setting with multi-exemplar pairing as shown in Figure~\ref{fig:abl_exemplars} (bottom), with an average accuracy of 55\% with ten times exemplar paired.
With multi-exemplar pairing in the conventional scenario, we observe from Figure~\ref{fig:abl_exemplars} (bottom) that as the system becomes increasingly plastic, this comes at the expense of stability, leading to heightened forgetting. We believe this is linked to the rapid parameter update and increased data flow during rehearsal. Although this facilitates quick learning, it simultaneously risks overwriting the weights responsible for encoding previously acquired knowledge. Additionally, in the conventional case, using equalization loss may not be entirely advantageous when the inherent randomness present in the data is adjusted. 

%%%%%%%%%%%%%%%%%
% Conclusion    %
%%%%%%%%%%%%%%%%
\section{Conclusion}
\label{sec:conclusion}
In this work, we focus on long-tailed online class incremental learning for image classification. We introduce a new two-stage method, DELTA, that decouples the learning of features from the classification task in the online setting using contrastive learning and an equalization loss. Additionally, we present early-stage work on multi-exemplar pairing in the LTOCL scenario. Our method shows significantly improved accuracy compared to existing OCL methods, showing a great potential to deploy online continual learning in real-life applications. 
For future work, we plan to explore representative exemplar selection within multi-exemplar pairing in the LTOCL setting to strike a balance between the stability and plasticity of the continual learner.

\clearpage
{
    \small
    \bibliographystyle{ieeenat_fullname}
    \bibliography{main}
}


\end{document}

%% file: preamble.tex
\usepackage[dvipsnames]{xcolor}